\documentclass[letterpaper, 10 pt, conference]{./style/IEEEconf}

\usepackage{amsmath}
\usepackage{mathtools}
\usepackage{amsfonts}
\usepackage{bm}
\usepackage{graphicx}
\usepackage{breqn} 
\usepackage{booktabs}
\usepackage{multirow}
\usepackage{cite}
\usepackage{url}

\usepackage{caption}
\usepackage{subcaption}
\DeclareMathOperator*{\argmin}{argmin}

\newcommand{\bmu}{\bm{\mu}}
\newcommand{\bw}{\bm{w}}
\newcommand{\bk}{\bm{k}}
\newcommand{\bK}{\bm{K}}
\newcommand{\bxi}{\bm{\xi}}
\newcommand{\bs}{\bm{s}}
\newcommand{\bPhi}{\bm{\Phi}}
\newcommand{\bSigma}{\bm{\Sigma}}
\newcommand{\bSigmainv}{\bm{\Sigma}^{-1}}
\newcommand{\bPsi}{\bm{\Psi}}

\graphicspath{{images-src/}{images-bin/}} 

\usepackage[textsize=scriptsize]{todonotes} 
\IEEEoverridecommandlockouts                             
\title{\LARGE \bf A Non-parametric Skill Representation with Soft Null Space Projectors  for Fast Generalization}

\author{Jo\~ao Silv\'erio$^1$ and Yanlong Huang$^2$
\thanks{$^1$ German Aerospace Center (DLR), Robotics and Mechatronics Center (RMC), M\"unchner Str. 20, 82234 We\ss ling, Germany. \texttt{joao.silverio@dlr.de}}
\thanks{$^2$ School of Computing, University of Leeds, Leeds
LS29JT, UK. \texttt{y.l.huang@leeds.ac.uk}}
}

\begin{document}

\maketitle

\begin{abstract}
Over the last two decades, the robotics community witnessed the emergence of 
various motion representations that have been used extensively, particularly in behavorial cloning, to compactly encode and generalize skills. Among these, probabilistic approaches have earned a relevant place,
owing to their encoding of variations, correlations and adaptability to new task conditions. Modulating such primitives, however, is often cumbersome due to the need for parameter re-optimization which frequently entails computationally costly operations. In this paper we derive a non-parametric movement primitive formulation that contains a null space projector. We show that such formulation allows for fast and efficient motion generation and adaptation with computational complexity $\mathcal{O}(n^2)$ without involving
matrix inversions, whose complexity is $\mathcal{O}(n^3)$. This is achieved by using the null space to track secondary targets, with a precision determined by the training dataset. 
Using a 2D example associated with time input we show that our non-parametric solution compares favourably with a state-of-the-art parametric approach. For demonstrated skills with high-dimensional inputs we show that it permits on-the-fly adaptation as well.
\end{abstract}

\section{Introduction}

Generalization is a fundamental challenge in machine learning, which in robot skill learning translates into modifying a behavior when new conditions arise. Since their inception with dynamic movement primitives (DMP) \cite{Ijspeert2002}, skill representations in robotics have targeted generalization. In DMPs this is done by design, by representing motion dynamics with a second-order linear system that ensures convergence to a pre-specified, alterable end-goal.
Alternative approaches using mixture models and hidden Markov models \cite{Calinon2007} are also endowed with strong generalization capabilities particularly under the framework of task-parameterized movement models \cite{Calinon2016, Huang18b}. Despite their popularity, these approaches lack analytical solutions to the adaptation problem, especially when via-points are not located at the start or end of the movement. In these situations, re-optimization of model parameters is required which can be prohibitively costly, e.g., reinforcement learning was used to optimize the parameters of DMP \cite{stulp2013robot} and GMM \cite{guenter2007reinforcement}. In \cite{saveriano2019merging} DMPs were combined in sequence to address the via-point issue, demanding several DMPs for a single skill, as opposed to traditional solutions requiring only one DMP.

A second line of approaches leverages probability theory to provide increased adaptation capabilities. Paraschos et~al. introduced probabilistic movement primitives (ProMP) \cite{Paraschos2013}. ProMP assumes that demonstrations of a task are approximated by a parametric trajectory amounting to a weighted 
sum of basis functions and uses Gaussian
conditioning to compute optimal weights to pass through new, desired via-points. 
ProMP has been extended to handle obstacle avoidance \cite{shyam2019improving, low2021prompt}, where in \cite{shyam2019improving} a signed distance between a robot arm and an obstacle was used similarly to 
CHOMP \cite{ratliff}, and in  \cite{low2021prompt} a random exploration and update scheme, initialized by ProMP weight distributions, was provided similarly to 
STOMP \cite{stomp}. 
In \cite{frank2021constrained} constrained ProMP was studied by formulating external constraints into a nonlinear optimization problem.
In \cite{Huang2019}, we introduced a non-parametric, kernelized movement primitives (KMP) formulation which uses kernel functions to mitigate the need to carefully define basis functions, a well-known issue in ProMP and related techniques relying on fixed basis functions. This is the case especially if the trajectory is not driven by time but by a multi-dimensional input, such as the position or posture of a human in the robot workspace, where the number of required basis functions grows quickly with the number of input variables \cite{Bishop2006}. We argue that formulations that allow to seamlessly encode dependencies on human motions, and not only time, have an important role to play in the near future as human-robot interaction gains traction \cite{maeda2017phase,cui2019environment,Silverio2018a,Silverio2018b}. 

Despite providing an elegant solution to adapt a (potentially multi-dimensional-input) trajectory to pass through regions that were not previously shown to the robot, the computational cost of 
dealing with such adaptations by KMP 
is non-negligible. This is because it relies on a kernel matrix that, depending on factors like the task and the features being encoded (e.g. robot end-effector poses, joint angles, velocities),
may be costly to invert. It should be noted that this is an intrinsic limitation of many non-parametric methods, including Gaussian processes \cite{Rasmussen2006}.

In this paper we address the fast adaptation problem by deriving a variant of KMP that contains a null space projector. We start from the original problem formulation, described in Section \ref{sec:kmp}, and modify the optimization objective such that a null space projector appears naturally (Section \ref{sec:ns-kmp}). Null space projectors play an important role in robot control \cite{Siciliano1991,Wampler1986,Mayorga1992,Deo1995,Dietrich2015} and we re-use some of those core ideas here in a different domain. Subsequently we show that the projector can be kernelized too, providing an elegant and computationally efficient solution to probabilistic trajectory adaptation (Section \ref{sec:kernel-nskmp}), including in problems with multi-dimensional inputs. We also discuss some of its properties (Section \ref{sec:properties}), particularly the fact that idempotence of the projector is not guaranteed and depends on the training data (thus we refer to it as a \textit{soft} null space projector). While on the surface this might appear to be a limitation, it allows for covariance-weighted exploration trivially. Finally we compare our approach to two baselines in 2D and 3D examples (Sections \ref{sec:experiment1}--\ref{sec:experiment2}).

\section{Kernelized Movement Primitives}
\label{sec:kmp}

In the KMP framework we are interested in estimating a model that predicts the value of an output variable $\bxi \in \mathbb{R}^D$ given observations of an input $\bs\in\mathbb{R}^I$ from a set of $M$ datapoints $\left\lbrace \bs_m, \bxi_m \right\rbrace^M_{m=1}$ typically, but not necessarily, collected from human demonstrations. Examples of $\bs$ include time or human hand positions, while $\bxi$ is often the robot end-effector pose and its derivatives \cite{Huang2019}. Similarly to ProMP, 
KMP assumes a parametric approximation of a trajectory by a weighted sum of basis functions, i.e., $\bxi(\bs)= \begin{bsmallmatrix} \bm{\phi}(\bs) & \dots & \bm{0} \\ \vdots & \ddots & \vdots \\ \bm{0} & \dots & \bm{\phi}(\bs) \end{bsmallmatrix}^\top\bw = \bPhi(\bs)^\top\bw$, where $\bw\in\mathbb{R}^{BD}$ are weights and $\bm{\phi}\in\mathbb{R}^B$ is a vector of basis functions whose values depend on $\bs$. Moreover, $\bw \sim \mathcal{N}\left(\bmu_w,\bSigma_w\right)$. ProMP finds $\bmu_w, \bSigma_w$ using maximum likelihood estimation (MLE) with samples of $\bw$ obtained from demonstrations. KMP departs from ProMP by further assuming that a \textit{reference trajectory distribution}
$\left\lbrace \bmu_n, \bSigma_n \right\rbrace^N_{n=1}$ 
is available to model $\mathcal{P}\left(\bxi|\bs_n\right)$, where $\bs_{n=1,\dots,N}$ are $N$ given inputs. It then finds $\bmu_w,\bSigma_w$ by minimizing the objective $J(\bmu_w,\bSigma_w)=\sum^N_{n=1}\!\mathrm{D}_{\mathrm{KL}}\!\bigg(\mathcal{N}\!\!\left(\bPhi(\bs_n)\!^\top\!\bmu_w, \bPhi(\bs_n)\!^\top\!\bSigma_w\bPhi(\bs_n)\right)\!\!\vert\vert\mathcal{N}\left(\bmu_n,\bSigma_n\right)\!\!\!\bigg)$.
In this paper we focus on finding the optimal values of $\bmu_w^*$, which is done by solving the \textit{regularized weighted least squares} problem (see \cite{Huang2019} for the detailed derivation):
\begin{align}
	\bmu^*_w \!= & \argmin_{\bmu_w} \sum^N_{n=1}\!\left(\bPhi\!\left(\bm{s}_n\right)^\top\!\!\bmu_w\!-\!\bmu_n\!\right)^\top \!\!\bSigmainv_n \!\left(\bPhi\!\left(\bm{s}_n\right)^\top\!\!\bmu_w\!-\!\bmu_n\!\right) \nonumber\\ & \hspace{5cm}+ \lambda\bmu_w^\top\bmu_w, \label{eq:orig_KMP}
\end{align} 
where $\lambda>0$ is a scalar.
Differentiating the objective in \eqref{eq:orig_KMP} and equating to zero results in
\begin{equation}
	\bmu^*_w = \underbrace{\bPhi\left(\bPhi^\top\bPhi + \lambda\bSigma\right)^{-1}}_{\left(\bPhi^\top\right)^\dagger}\bmu, \label{eq:orig_KMP_sol}
\end{equation}
where we have  
\begin{gather}
    \bPhi = \left[\bPhi\left(\bm{s}_1\right) \ldots \bPhi\left(\bm{s}_N\right) \right], \quad\bmu =  \left[\bmu_1^\top \ldots \bmu_N^\top\right] \nonumber^\top, \nonumber \\ 
    \bSigma = \mathrm{blockdiag}\left(\bSigma_1,\ldots,\bSigma_N\right).\label{eq:kmp:variable}
\end{gather}
The underscored term in Eq. \eqref{eq:orig_KMP_sol} $\left(\bPhi^\top\right)^\dagger$ can be interpreted as a regularized right pseudo-inverse of $\bPhi^\top$ with regularization term $\lambda\bSigma$.

For a test input $\bs^*$, the expectation of $\bm{\xi}(\bs^*)$ under the assumption of a parametric trajectory is given by ${\mathbb{E}\left[\bxi(\bs^*)\right]= \bPhi(\bs^*)^\top\bmu_w^*}$. By replacing \eqref{eq:orig_KMP_sol} and applying the  \textit{kernel trick} \cite{Bishop2006}, we obtain
\begin{align}
	\mathbb{E}\left[\bxi(\bs^*)\right] \!=\! \bPhi(\bs^*)^\top\bmu_w^* & = \bPhi(\bs^*)^\top \bPhi\left(\bPhi^\top\bPhi + \lambda\bSigma\right)^{-1}\!\!\bmu \\
	& = \bk^*\left(\bK + \lambda\bSigma\right)^{-1}\bmu, \label{eq:kmp_expectation_orig}
\end{align}
where $\bk^*=\left[\bk(\bs^*,\bs_1),\ldots,\bk(\bs^*,\bs_N)\right]$, ${\bK = \begin{bsmallmatrix} \bk(\bs_1,\bs_1) & \dots & \bk(\bs_1,\bs_N) \\ \vdots & \ddots & \vdots \\ \bk(\bs_N,\bs_1) & \dots & \bk(\bs_N,\bs_N) \end{bsmallmatrix}}$, $\bk(\bs_i,\bs_j)=k(\bs_i,\bs_j)\bm{I}$ and $k(\bs_i,\bs_j) = \bm{\phi}(\bs_i)^\top\bm{\phi}(\bs_j)$ is a kernel function from hereon assumed to be the squared-exponential $k(\bs_i,\bs_j) = \mathrm{exp}(-\frac{1}{l^2}\vert\vert \bs_i-\bs_j\vert\vert^2)$.

From \eqref{eq:orig_KMP}, \eqref{eq:kmp_expectation_orig} it follows that if, for a certain $\bmu_n$, the covariance $\bSigma_n$ is small, the expectation at $\bs_n$ will be close to $\bmu_n$. This provides a principled way for trajectory modulation. Indeed, if, for a new input $\bar{\bs}$, one wants to ensure that the expected trajectory passes through a desired $\bar{\bmu}$ it suffices to manually add the pair $\{\bar{\bmu},\bar{\bSigma}\}$ to the reference distribution 
provided that $\bar{\bSigma}$ is small enough. While this solves the adaptation problem trivially, it comes at the cost of having to invert the term $\bK+\lambda\bSigma$ every time a new point is added. This hinders the applicability of KMP in scenarios that require a fast evaluation of variations of the demonstrated trajectories. We propose to alleviate this issue by formulating trajectory modulation as acting on the null space of the movement primitive.

\section{A null space formulation of KMP}

\subsection{KMP with null space as the solution to a least squares problem}
\label{sec:ns-kmp}

We take inspiration from the robotics literature on null space projectors \cite{Deo1995,Mayorga1992,Wampler1986} and extend the original problem \eqref{eq:orig_KMP} 
 with an additional cost term to keep the solution close to a desired one denoted by $\hat{\bw}$, i.e.,
\begin{align}
	\bmu^*_w = \argmin_{\bmu_w} \left(\bPhi^\top \bmu_w-\bmu\right)^\top \bSigmainv \left(\bPhi^\top\bmu_w-\bmu\right) + \alpha\bmu_w^\top\bmu_w \nonumber \\ \underbrace{\hspace{2.5cm}+ \beta\left(\bmu_w-\hat{\bw}\right)^\top\left(\bmu_w-\hat{\bw}\right)}_{J(\bmu_w)}.
\end{align}
The solution can be obtained similarly to \eqref{eq:orig_KMP} by solving $\frac{\partial J(\bmu_w)}{\partial\bmu_w}=0$, leading to
%
%
%
\begin{equation}
	\bmu^*_w \!=\! \left(\bPhi\bSigmainv\bPhi^\top \!+\! \left(\alpha + \beta\right) \bm{I}\right)^{-1} \! \left(\bPhi\bSigmainv\bmu + \beta\hat{\bw}\right).\label{eq:null_sol_prelim_0}
\end{equation}
By assuming $\lambda = \alpha + \beta$,  \eqref{eq:null_sol_prelim_0} is re-written as
%
%
%
\begin{equation}
	\bmu^*_w = \bPhi\left(\bPhi^\top\bPhi \!+\! \lambda\bSigma\right)^{-1}\bmu \!+\! \beta\left(\bPhi\bSigmainv\bPhi^\top \!+\! \lambda\bm{I}\right)^{-1}\!\hat{\bw}.\label{eq:null_sol_prelim}
\end{equation}
Moreover, by using the Woodbury identity\footnote{$\left(\bm{A}+\bm{C}\bm{B}\bm{C}^\top\right)^{-1} = \bm{A}^{-1} + \bm{A}^{-1}\bm{C}\left(\bm{B}^{-1}+\bm{C}^\top\bm{A}^{-1}\bm{C}\right)^{-1}\bm{C}^\top\bm{A}^{-1}$} the inverse in the second term of \eqref{eq:null_sol_prelim} takes the form
\begin{equation}\!
	\left(\bPhi\bSigmainv\bPhi^\top\! \!+\! \lambda\bm{I}\right)^{-1} = \frac{1}{\lambda}\left[\bm{I} - \bPhi\left(\bPhi^\top\bPhi \!+\! \lambda\bSigma\right)^{-1}\bPhi^\top\right]
\end{equation}
which plugged back into \eqref{eq:null_sol_prelim} results in
\begin{equation}
	\bmu^*_w \!=\! \underbrace{\bPhi\left(\bPhi^\top\!\bPhi \!+\! \lambda\bSigma\right)^{-1}}_{\left(\bPhi^\top\right)^\dagger}\!\!\bmu + \frac{\beta}{\lambda}\!\!\left[\bm{I} \!-\! \underbrace{\bPhi\left(\bPhi^\top\!\bPhi \!+\! \lambda\bSigma\right)^{-1}}_{\left(\bPhi^\top\right)^\dagger}\bPhi^\top\right]\!\!\hat{\bw}.\label{eq:KMP_with_null_space_sol}
\end{equation}
Equation \eqref{eq:KMP_with_null_space_sol} corresponds to a typical least squares solution with null space where for a linear system $\bm{Y}=\bm{X}\bm{A}$ we have as solution ${\hat{\bm{Y}}=\hat{\bm{X}}\bm{A}^*=\hat{\bm{X}}\left(\bm{X}^\dagger \bm{Y} + \bm{N}\bm{v}\right)}$ where $\bm{v}$ satisfies a secondary goal and $\bm{N} = \bm{I} - \bm{X}^\dagger\bm{X}$.

For the remainder of the discussion we will assume, without loss of generality, $\frac{\beta}{\lambda}=1$. For reference, this assumption is common in robot control when damped least squares and null space projectors are combined, see \cite{Deo1995,Mayorga1992}.

\subsection{Kernelizing the null space solution}
\label{sec:kernel-nskmp}

As for the original solution of KMP, the optimal solution $\bmu_w^*$ is applied to a new test point to predict the expectation of the output as
\begin{dmath}
	\mathbb{E}\left[\bxi(\bm{s}^*)\right] 
	= \bPhi(\bm{s}^*)^\top\bmu_w^*\\
	= \bPhi^\top(\bm{s}^*)\left( \bPhi\left(\bPhi^\top\bPhi + \lambda\bSigma\right)^{-1}\bmu + \left[\bm{I} - \bPhi\left(\bPhi^\top\bPhi + \lambda\bSigma\right)^{-1} \bPhi^\top\right]\hat{\bw} \right)\\
	= \bPhi(\bm{s}^*)^\top\bPhi\left(\bPhi^\top\bPhi + \lambda\bSigma\right)^{-1}\!\!\bmu \!+\! \left[\bPhi(\bm{s}^*)^\top - \bPhi(\bm{s}^*)^\top\bPhi\left(\bPhi^\top\bPhi + \lambda\bSigma\right)^{-1} \bPhi^\top\right]\hat{\bw}.
\end{dmath}
Using the \textit{kernel trick} we can write
\begin{dmath}
	\mathbb{E}\left[\bxi(\bm{s}^*)\right] = \bk^*\left(\bK + \lambda\bSigma\right)^{-1}\bmu + \left[\bPhi(\bm{s}^*)^\top - \bk^*\left(\bK + \lambda\bSigma\right)^{-1} \bPhi^\top\right]\hat{\bw}.\label{eq:kernelized_sol_prelim}
\end{dmath}
For the sake of the derivation, we convert the null space desired weights $\hat{\bw}$ into a desired set of points in trajectory space. From the assumption of a parametric trajectory $\bm{\xi}(\bs) = \bPhi\!(\bs)^\top\bw$ we define an arbitrary number $P$ of desired secondary target points as ${\hat{\bm{\xi}}_1 = \bPhi(\bm{s}_{p=1})^\top \hat{\bw},\>\> \ldots,  \hat{\bm{\xi}}_P = \bPhi(\bm{s}_{p=P})^\top \hat{\bw}}$
or, in matrix form,
\begin{equation}
	\hat{\bm{\xi}} = \hat{\bPhi}^\top \hat{\bw}.
\end{equation}
Next we approximate the optimal values of $\hat{\bw}$ given the desired secondary target points $\hat{\bm{\xi}}$ using the right pseudo-inverse of $\hat{\bPhi}^\top$, i.e.
\begin{equation}
	\hat{\bw} = \hat{\bPhi}\left(\hat{\bPhi}^\top \hat{\bPhi}\right)^{-1}\hat{\bm{\xi}}. \label{eq:weights_to_points}
\end{equation}
Plugging \eqref{eq:weights_to_points} back in \eqref{eq:kernelized_sol_prelim} yields
\begin{dmath}
	\mathbb{E}[\bxi(\bm{s}^*)] = \bk^*\left(\bK + \lambda\bSigma\right)^{-1}\bmu + \left[\bPhi(\bm{s}^*)^\top - \bk^*\left(\bK + \lambda\bSigma\right)^{-1} \bPhi^\top\right]\hat{\bPhi}\left( \hat{\bPhi}^\top\hat{\bPhi}\right)^{-1}\hat{\bm{\xi}}\nonumber \\
 	= \bk^*\left(\bK + \lambda\bSigma\right)^{-1}\bmu + \left[\bPhi(\bm{s}^*)^\top\hat{\bPhi} - \bk^*\left(\bK + \lambda\bSigma\right)^{-1} \bPhi^\top\hat{\bPhi}\right]\left( \hat{\bPhi}^\top \hat{\bPhi}\right)^{-1}\hat{\bm{\xi}}\nonumber
 	\\
	 = \bk^*\left(\bK + \lambda\bSigma\right)^{-1}\bmu + \left[\hat{\bk}^* - \bk^*\left(\bK + \lambda\bSigma\right)^{-1}\hat{\bK}\right]\left(\underline{\bK}\right)^{-1}\hat{\bm{\xi}}\nonumber
	\\
	= \bk^*\bPsi\bmu + \left[\hat{\bk}^* - \bk^*\bPsi\hat{\bK}\right]\left(\underline{\bK}\right)^{-1}\hat{\bm{\xi}}. \label{eq:kmp_null_final_0}
\end{dmath}
with $\hat{\bk}^* = \bPhi(\bm{s}^*)^\top\hat{\bPhi}$, $\hat{\bK} = \bPhi^\top\hat{\bPhi}$, $\underline{\bK} = \hat{\bPhi}^\top\hat{\bPhi}$ and $\bPsi = \left(\bK + \lambda\bSigma\right)^{-1}$.

Finally, for a single desired secondary target and a squared-exponential kernel we have $\underline{\bK}=\bm{I}$ and \eqref{eq:kmp_null_final_0} becomes\footnote{To contrast our solution with the `classical' KMP \cite{Huang2019}, throughout the rest of the paper we refer to \eqref{eq:kmp_null_final} as the \textit{null space KMP} (NS-KMP). In addition, we refer to $\hat{\bxi}$ as the \textit{null space reference}.}
\begin{equation}
	\mathbb{E}[\bxi(\bm{s}^*)] = \bk^*\bPsi\bmu + \left[\hat{\bk}^* - \bk^*\bPsi\hat{\bK}\right]\hat{\bm{\xi}}.\label{eq:kmp_null_final}
\end{equation}
Notice that \eqref{eq:kmp_null_final} corresponds to \eqref{eq:kmp_expectation_orig} with the additional null space term $\left[\hat{\bk}^* - \bk^*\bPsi\hat{\bK}\right]\hat{\bm{\xi}}$. Equation \eqref{eq:kmp_null_final} thus modulates the expectation \eqref{eq:kmp_expectation_orig} in a computationally efficient manner. We leverage this property by modifying $\hat{\bxi}$ such that variations of the original trajectory are achieved, similarly to what is done in robot null space exploration \cite{park2008movement, Huang2018,  schneider2010robot}.
However, unlike the null spaces found in robot control which are often strict (see \cite{Deo1995} for an overview) the KMP null space possesses a set of properties derived from its probabilistic foundations that dictate how $\hat{\bxi}$ modulates \eqref{eq:kmp_expectation_orig}.

\section{Properties of the kernelized null space projector}
\label{sec:properties}

\subsection{Idempotence and `soft' null space}

A general property of null space projectors is idempotence, i.e. for a null space projector $\bm{N}$ we have $\bm{N}\bm{N} = \bm{N}$ \cite{Dietrich2015}. The projector derived in Section \ref{sec:ns-kmp} is not guaranteed to have this property since, generally, ${\left(\bm{I}-\left(\bPhi^\top\right)^\dagger\bPhi^\top\right)\left(\bm{I}-\left(\bPhi^\top\right)^\dagger\bPhi^\top\right)\neq \left(\bm{I}-\left(\bPhi^\top\right)^\dagger\bPhi^\top\right)}$. It should be noted, however, that as $\lambda\bSigma \rightarrow \bm{0}$ idempotence is fulfilled, which can be verified trivially. In practice, this means that the larger $\bSigma$, the more the null space reference is allowed to deform the original trajectory. For this reason we refer to it as a \textit{soft} null space projector. In robotics, this is known to happen when computing null space projectors using the regularized pseudo-inverse of the Jacobian matrix. Indeed, in those cases, the null space tasks are known to affect the precision of the main one \cite{Deo1995}.

\begin{figure}
\centering
\includegraphics[width=0.95\columnwidth, trim={0, 0.2cm, 0, 0}, clip]{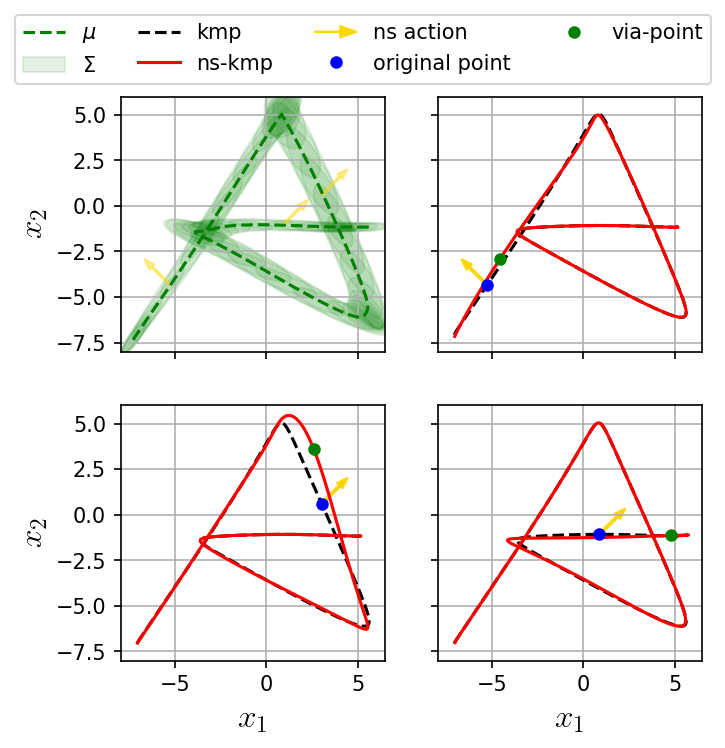}
\caption{Effect of the null space projector in modulating the original trajectory. \textit{Top-left:} reference distribution with mean shown as a dashed curve and  covariance as ellipses. \textit{Top-right and bottom:} Modulation examples at three different points. A null space reference with the same magnitude across all plots is applied in the direction of the yellow arrow at the blue point. The non-modulated KMP trajectory (dashed) changes its shape (red) depending on the covariance at the point where the reference is applied. The green point is the via-point one would need to define to achieve the same modulation using classical KMP adaptation (see Section \ref{sec:ns-equiv}).}
\vspace{-0.4cm}
\label{fig:ns-effect}
\end{figure}

\subsection{Covariance-weighted null space modulation}

Unlike the robot control case, where regularization typically appears as a diagonal matrix, $\bSigma_n$ are full covariance matrices. As a consequence, a null space reference generally does not affect all dimensions of the trajectory equally. The covariance at a given point influences both the magnitude and the directions along which a null space reference can modulate the trajectory. Figure \ref{fig:ns-effect} illustrates this property. The top-left image shows a trajectory distribution in 2D computed from examples of drawing a letter `A' \cite{Calinon2019}. That distribution was used as a reference trajectory distribution in a KMP. Notice how the covariance changes throughout the trajectory. In the top-right and bottom images, we show the effect of applying a null space reference (yellow arrow), with the same magnitude,  at different points along the trajectory. The dashed line is computed from \eqref{eq:kmp_expectation_orig} and the solid, red one from \eqref{eq:kmp_null_final}. The covariance in the null space projector constrains the modulated trajectory to deform less (more) along the directions where the variance is lower (higher). Additionally the modulation is applied along the axes of $\bSigma_n$. 
This provides a principled way to use the variations in the original dataset to design optimal exploration strategies. Moreover, unlike typical exploration strategies that require the re-computation of model parameters or computationally costly operations to include new points in the trajectory, here adding new points is done inexpensively with \eqref{eq:kmp_null_final}. Indeed, the parameter $\bPsi$, that contains a potentially large matrix inverse operation, is computed only once from the reference trajectory distribution.

\begin{figure}
\centering
\includegraphics[width=0.95\columnwidth, trim={0, 0.2cm, 0, 0}, clip]{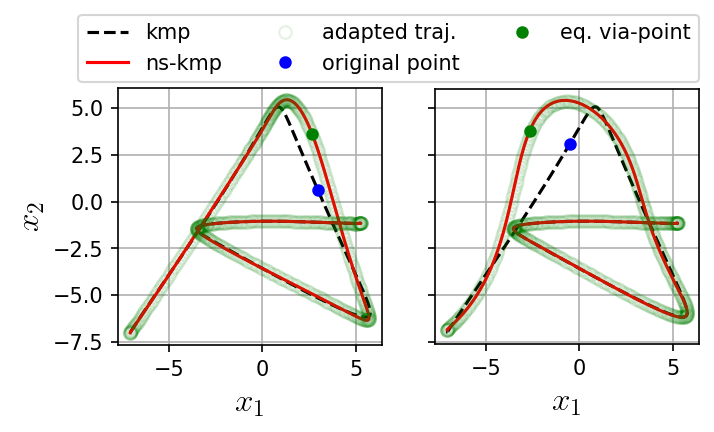}
\caption{Equivalence between null space modulation and via-point adaptation. A null space reference $\hat{\bxi}$ applied at input $\hat{\bs}$ modulates the original KMP (dashed), computed from \eqref{eq:kmp_expectation_orig}, generating the red line, computed from  \eqref{eq:kmp_null_final}. In the dashed line, $\hat{\bs}$ resulted in the blue point. In the red line $\hat{\bs}$ resulted in the green point. By treating the latter as a via-point (see Section \ref{sec:kmp}), classical KMP via-point adaptation (green circled line) produces the same result.}
\vspace{-0.4cm}
\label{fig:ns-via-equiv}
\end{figure}

\subsection{Equivalence to classical KMP adaptation}
\label{sec:ns-equiv}

An important final property of NS-KMP is that null space modulation is equivalent to classical KMP modulation \cite{Huang2019} given the appropriate via-point. For the sake of the explanation let us consider an input $\hat{\bs}$ where a null space reference $\hat{\bxi}$ was applied, resulting in an observation $\bxi(\hat{\bs}) \neq \hat{\bxi}$ after the null space projector. We have concluded empirically that by treating the pair $\{\hat{\bs},\bxi(\hat{\bs})\}$ as a via-point in the classical KMP framework \cite{Huang2019}, and adding it to the reference trajectory distribution with a low covariance, generates the same trajectory as the null-space-generated one. Figure \ref{fig:ns-via-equiv} illustrates this property. This result suggests that NS-KMP is, indeed, a computationally-efficient tool for via-point-based motion adaptation.

\section{Experiment I}
\label{sec:experiment1}

We first report results on the re-planning of a 2D trajectory, learned from demonstrations, after an obstacle appears. We tackle this problem using null space exploration (NS-KMP) or by adding via-points (KMP, ProMP) to the demonstrated trajectory distribution. 
We compare the time performance of KMP, NS-KMP and ProMP.

\subsection{Setup}

We use the handwritten letter dataset available in \cite{Calinon2019}, particularly the dataset for letter `A', to generate a trajectory distribution from demonstrations. It contains 10 demonstrations with time $t$ and 2D position $(x_1,x_2)$, which we use to train a mixture model with 8 components and retrieve a reference trajectory distribution $\{\bmu_n,\bSigma_n\}_{n=1}^{100}$ using Gaussian mixture regression. Moreover $\bs = t$ and $\bxi=\left[x_1 \> x_2\right]^\top$. In both KMP and NS-KMP implementations we use hyperparameters $\lambda = 0.1$, $l = 0.125$ while for ProMP we use 20 basis functions. The implementation was done in Python 3.8 running on an 8-core 11-Gen Intel Core i7-11850H @ 2.50GHz processor.

We use the learned models to simulate a robot planning a trajectory within a finite-horizon. At time $t$, the robot uses KMP, NS-KMP or ProMP to compute a trajectory for the horizon $\{t,\ldots,T\}$. We use $T=1$ with $10ms$ increments between time steps. At a randomly chosen time instant in the horizon, an obstacle (a circle with radius 0.5) with position uniformly sampled from $\{\bmu_n\}_{n=1}^N$ appears (only once) along the planned path. The robot then tests alternative paths looking for one that does not collide with the obstacle. In NS-KMP, this is done by sampling values of input $\hat{\bs}$ and null space reference $\hat{\bxi}$ and computing the corresponding new trajectory using \eqref{eq:kmp_null_final}. The sampling interval for the input is $[t,\ldots,\mathrm{min}(t+0.2,T-t)]$ and $[-1000,1000]$ for the two dimensions of $\hat{\bxi}$ (we discuss this range of values in Section \ref{sec:discussion}). In KMP and ProMP, finding alternative paths is done by sampling via-points in the neighborhood of the obstacle and inputs in the same range as for NS-KMP. To efficiently perform the exploration, including sampling null space references and via-points we used TPE \cite{Bergstra2011}, particularly the implementation available in the Optuna package \cite{Akiba2019}. We define a simple objective function to minimize $f(\bm{X}_{pred})=c_{collision}(\bm{X}_{pred})+c_{cont}(\bm{X}_{pred})$ where $\bm{X}_{pred}$ is the sequence of 2D points in the planned trajectory and $c_{collision}$ and $c_{cont}$ are collision and continuity terms that receive values of 1000 if a collision happens in $\bm{X}_{pred}$ or the norm between any two consecutive points in $\bm{X}_{pred}$ exceeds the radius of the obstacle and 0 otherwise. The optimizer returns the first solution with zero cost.

\begin{figure}
\centering
\begin{subfigure}[b]{\columnwidth}
	\centering
	\includegraphics[width=0.49\columnwidth, trim={0, 0.2cm, 0, 0}, clip]{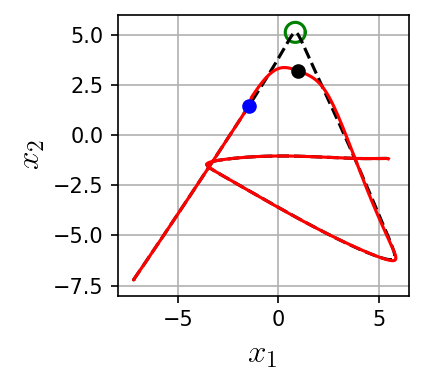}
	\includegraphics[width=0.49\columnwidth, trim={0, 0.2cm, 0, 0}, clip]{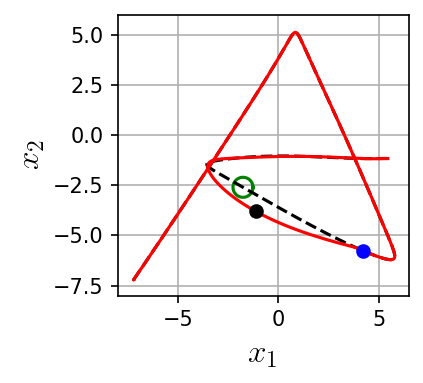}
	\caption{Modulations with one via-point.}
	\label{fig:ns-obs-result}
\end{subfigure}
\begin{subfigure}[b]{\columnwidth}
	\centering
	\includegraphics[width=0.49\columnwidth, trim={0, 0.2cm, 0, 0}, clip]{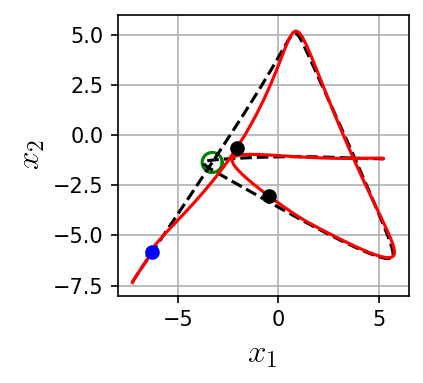}
	\includegraphics[width=0.49\columnwidth, trim={0, 0.2cm, 0, 0}, clip]{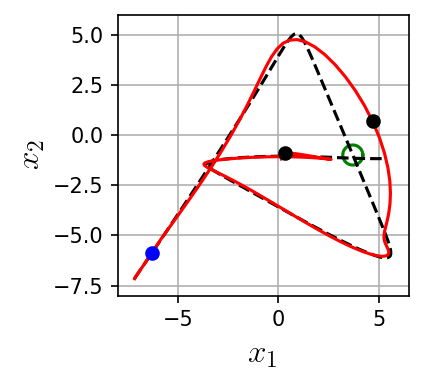}
	\caption{Modulations with two via-points.}
	\label{fig:ns-2obs-result}
\end{subfigure}
\caption{Examples of trajectories obtained using NS-KMP for different numbers of via-points. The blue point show the robot position when the obstacle (green circle) appears while the black point depicts the via-point(s) discovered by NS-KMP, which result in an adapted trajectory (red). The dashed line is the nominal trajectory encoded in the KMP.}
\vspace{-0.4cm}
\end{figure}

\subsection{Results using one via-point}
\label{sec:experiment1_A}

Figure \ref{fig:ns-obs-result} shows example trajectories obtained from NS-KMP. We observe that the obstacle is avoided correctly in different segments of the trajectory. Table \ref{tab:time_results_exp1} summarizes the time required by KMP, NS-KMP and ProMP to predict a trajectory from one null space reference or via-point (first row) and to find a solution to avoid the obstacle (second row). Each entry corresponds to the mean and standard deviation of 40 points. In the case of NS-KMP one \textit{prediction} corresponds to computing $\hat{\bk}^*$, $\hat{\bK}$ and \eqref{eq:kmp_null_final}. For KMP it is the time of computing new matrices $\bK$, $\bk^*$ and \eqref{eq:kmp_expectation_orig}, including the inversion. For ProMP it is the time to evaluate the basis functions at the new input, recompute the weight distribution and use it to compute a new trajectory (see \cite{Paraschos2013}). Note that in this experiment we did not allow obstacles to appear at points where the trajectory intersects with itself (see these cases in Fig. \ref{fig:ns-2obs-result}) since they require two via-points. The \textit{prediction} times show that the parametric model ProMP was the fastest, closely followed by our non-parametric NS-KMP. Moreover, we observed comparable times between NS-KMP and ProMP to find a good trajectory (second row, \textit{1 via-point (s)}), with KMP (first column) taking two orders of magnitude longer.

\subsection{Results for collision avoidance with two via-points}

Finally we investigated modulation by more than one via-point. For this we focused on the cases where the obstacle appears at places where the trajectory intersects with itself (Fig. \ref{fig:ns-2obs-result}). Given the high computational cost of KMP shown in \ref{sec:experiment1_A} we compare NS-KMP and ProMP only. Figure \ref{fig:ns-2obs-result} shows that NS-KMP is capable of avoiding obstacles by modulating the planned trajectory with two via-points.  As seen in Table \ref{tab:time_results_exp1}, last row, NS-KMP and ProMP achieved times within the same order of magnitude, with a slight disadvantage to ProMP.

\begin{table}[t]
\centering
\caption{Time to find solutions in 2D trajectory modulation.}
\begin{tabular}{@{}p{1.8cm}lll@{}}
\toprule
                    & \textbf{KMP} & \textbf{NS-KMP (ours)} & \textbf{ProMP} \\ \midrule
                    Prediction (ms) & $67.4 \pm 4.6$ & $1.12 \pm 0.33$ & $0.24 \pm 0.02$ \\ \midrule
1 via-point (s) & $1.62 \pm 1.14$ & $0.05 \pm 0.04$ & $0.02 \pm 0.04$\\ \midrule
2 via-point (s) & - & $1.70 \pm 1.40$ & $5.63 \pm 6.20$ \\\bottomrule
\end{tabular}
\label{tab:time_results_exp1}
\end{table}

\section{Experiment II}
\label{sec:experiment2}

In a second experiment we investigate the ability of NS-KMP to modulate trajectories generated from multi-dimensional inputs. 
We consider a handover task where the end-effector position of one robot is computed from the one of another robot and adapted after an obstacle appears.

\subsection{Setup}

In order to train a NS-KMP we used the dataset collected from handover demonstrations in \cite{Silverio2019}, with 5 demonstrations in the form of human hand and robot end-effector positions  $\bm{x}_H$  and $\bm{x}_R$.
The NS-KMP thus uses $\bs = \bm{x}_H$ and $\bxi = \bm{x}_R$. In order to simulate the human-robot handover task, the data was rescaled to fit the robot workspaces in the \textit{pybullet} setup depicted in Fig. \ref{fig:handovers}. As illustrated in Fig.~\ref{fig:handovers}, the left robot plays the role that the human played in the demonstrations by picking up an object (yellow box) and handing it over to the robot on the right. The latter starts with a planned motion (black line) computed using as input the trajectory planned by the left robot (unavailable in a real scenario with a human, only shown for illustration). In each trial the motion is interrupted by a cube-shaped obstacle with an axial diagonal of $10cm$ that appears randomly (both in time and location) once along the black line. All the NS-KMP parameters were the same as in \ref{sec:experiment1}. The optimization framework was also the same, however $\hat{\bxi}$ was sampled from the range $[-2000,2000]$ and the inputs (i.e.  the left robot end-effector position) were sampled uniformly from the ones in the reference trajectory distribution. As in \cite{Huang2019},  we do not use ProMP as baseline here due to the difficulty in parameterizing basis functions for multi-dimensional inputs. Instead, we consider the optimization of Gaussian components in GMM as a baseline \cite{guenter2007reinforcement}. Specifically, we optimize the mean of one randomly chosen Gaussian until the collision cost is zero. 
We do not evaluate classical KMP in this section as the arguments that caused it to underperform in \ref{sec:experiment1_A} are still valid (see also the analysis of computational complexity in Section~\ref{subsec:comp:ana}).

\subsection{Results}

\begin{figure}
\centering
\includegraphics[width=0.49\columnwidth,trim={5cm 5cm 5cm 5cm},clip]{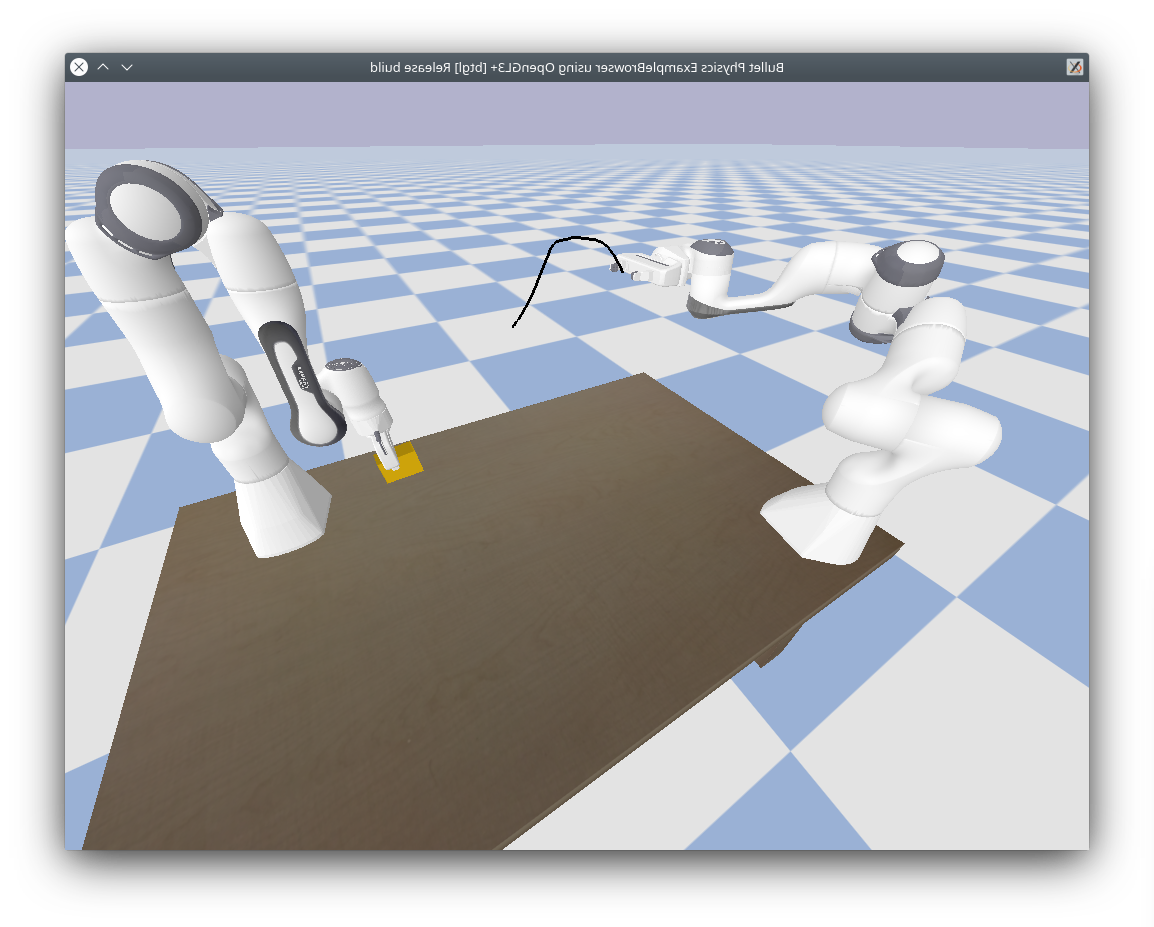}
\includegraphics[width=0.49\columnwidth,trim={5cm 5cm 5cm 5cm},clip]{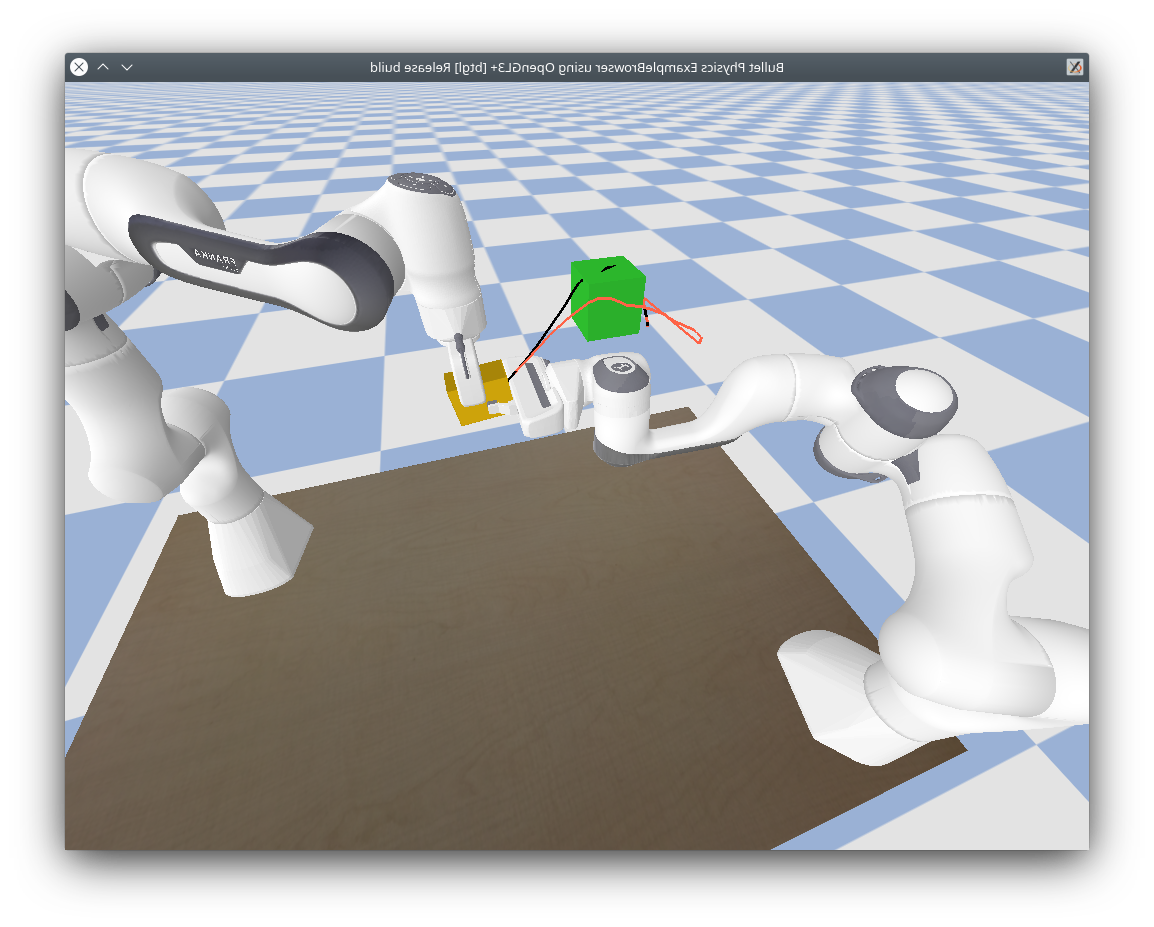}\\ \vspace{0.1cm}
\includegraphics[width=0.49\columnwidth,trim={5cm 5cm 5cm 5cm},clip]{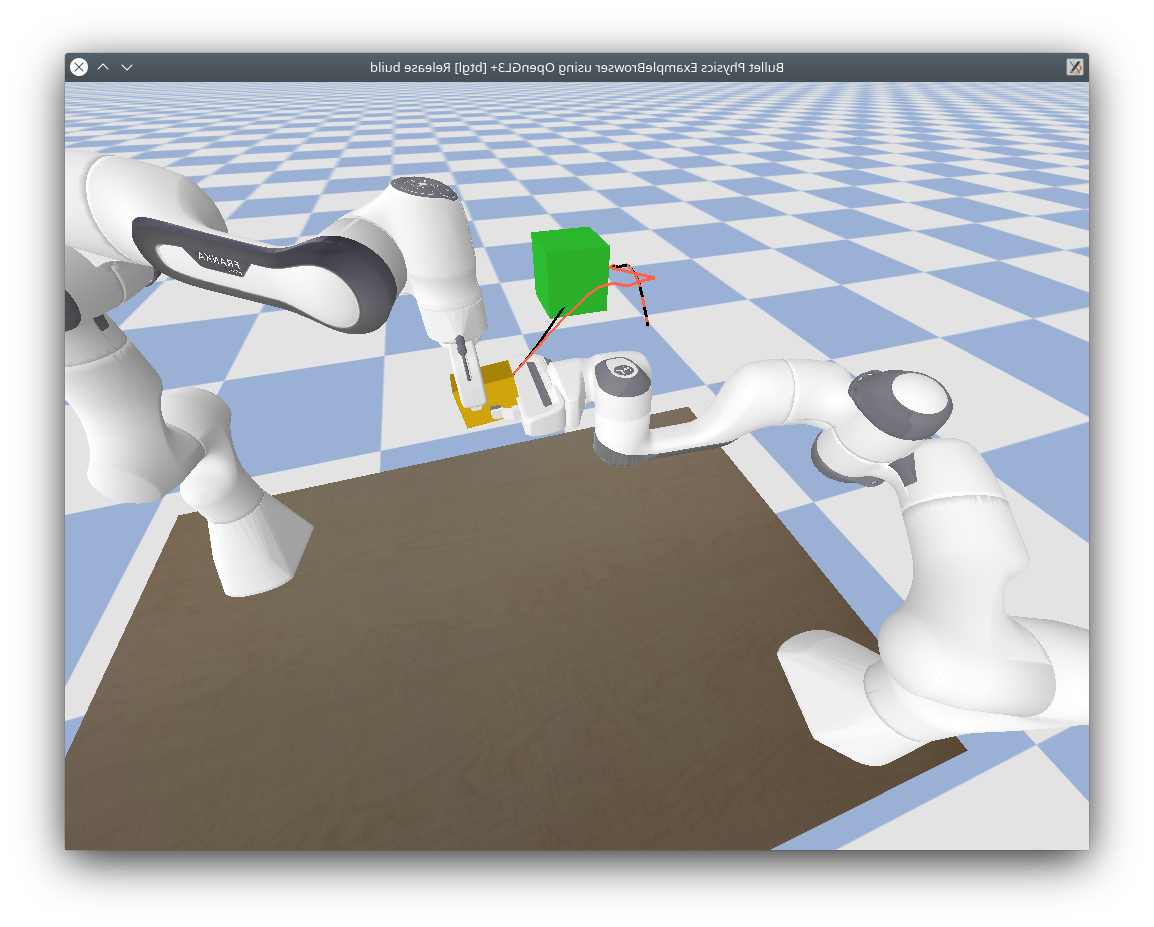}
\includegraphics[width=0.49\columnwidth,trim={5cm 5cm 5cm 5cm},clip]{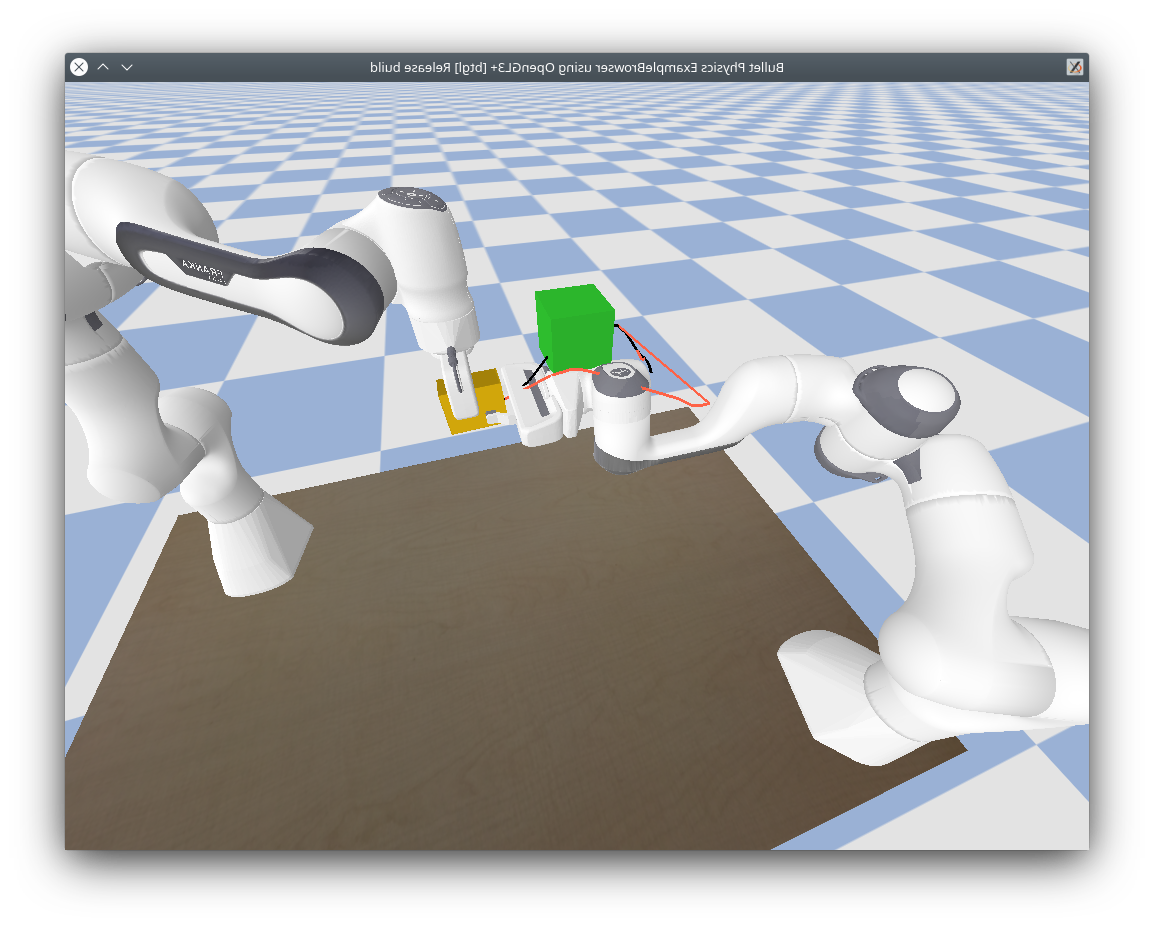}
\caption{\textit{Top-left:} Start of the handover task. \textit{Top-right and bottom:} Reproductions in simulation. The right robot motion is governed by the position of the left robot end-effector. For illustration, the black line shows the nominal trajectory of the right arm based on the planned motion of the left arm. An obstacle (green) appears during motion. Using NS-KMP the right arm re-plans its motion resulting in the trajectory in orange.}
\vspace{-0.4cm}
\label{fig:handovers}
\end{figure}

In 20 trials, the average time to compute a solution with NS-KMP after the obstacle appears was $0.16s \pm 0.13$. Fig. \ref{fig:handovers}, \textit{top-right} and \textit{bottom}, shows original and adapted trajectories. The GMM baseline yielded $0.616s \pm 0.571$.

\section{Discussion}
\label{sec:discussion}

\subsection{Analysis of experimental results}

The results in Section \ref{sec:experiment1} revealed that our solution, despite not outperforming ProMP ($\approx 5\times$ faster at predicting), computes trajectories given a new desired point at the millisecond level  (Table \ref{tab:time_results_exp1}, first row) with a non-optimized Python implementation.
We believe this to be a noteworthy result for a non-parametric approach. Moreover, ProMP and NS-KMP performed at a similar level in finding a 1-via-point solution. However, ProMP is only $\approx 2.5 \times$ faster, against $5\times$ for the individual prediction, suggesting that ProMP needs to try more via-points than NS-KMP to find a solution. This could be related to our choice of sampling via-points for ProMP around the obstacle, which may not be as efficient as the covariance-weighted exploration by NS-KMP. 
The reason for such choice is that we observed that the trajectories generated by ProMP often had arbitrary shapes (seriously violating the structure of the training data) if the search space of the via-point is not bounded or is large. The same reasoning applies to the 2-via-point case where NS-KMP took less time to find successful via-points. 

Adaptation of learned multi-dimensional-input skills was a novelty introduced by KMP \cite{Huang2019} and the results in Section \ref{sec:experiment2} show that NS-KMP provides a fast solution to adapting this type of skills.
The GMM baseline was slower and the resulting trajectories were prone to distortions since only one Gaussian was being optimized at a time. The presented times for GMM can thus be seen as a lower bound since optimizing further parameters would require higher amounts of time.
Despite the successful adaptation with NS-KMP we verified that, for our choice of hyperparameters, the time to compute a successful trajectory increases when the object appears closer to the handover location due to the decrease in variance (the null space `hardens'). 

Some additional practical aspects of NS-KMP are worth discussing. As described in \ref{sec:experiment1}--\ref{sec:experiment2} the magnitude of $\hat{\bxi}$ could be as high as 1000 or 2000 depending on the problem. Due to the probabilistic nature of the \textit{soft} null space projector, one does not know \textit{a priori} the required magnitude of $\hat{\bxi}$. It should thus be treated as a hyperparameter to be tuned depending on the problem. Additionally, while covariance-weighted null space exploration seems to alleviate the need to manually define exploration strategies, in some scenarios one might need to violate the priors imposed by the data, e.g. if the object has a much larger size. In other words it may be desirable to `soften' the null space. This can be achieved by manually increasing $\lambda\bSigma$, particularly in the neighborhood of the concerned region. 

\subsection{Analysis of computational complexity}
\label{subsec:comp:ana}
Classical KMP \cite{Huang2019} predicts the expectation of $\bxi(\bs^*)$ using \eqref{eq:kmp_expectation_orig}.
When a new desired point described by the mean $\bar\bmu$ and covariance $\bar\bSigma$ is needed, the reference trajectory distribution $\left\lbrace \bmu_n, \bSigma_n \right\rbrace^N_{n=1}$ is concatenated with $\{\bar\bmu$,$\bar\bSigma\}$, leading to an extended reference trajectory with length $N+1$. Consequently, KMP needs to recompute the term $\left(\bK + \lambda\bSigma\right)^{-1}$ since the updated $\bSigma$ includes $\bar\bSigma$, see \eqref{eq:kmp:variable}. Note that the dimension of $\bK$ and $\bSigma$ is $D(N+1) \times D(N+1)$ and therefore the computational complexity of classical KMP is $\mathcal{O}(D^3(N+1)^3)$. Recall that in \eqref{eq:kmp_null_final} we only search in the null space, where $\bK$ remains the same and therefore $\bPsi$ can be computed beforehand in an offline fashion. Therefore, the computation complexity of \eqref{eq:kmp_null_final} is $\mathcal{O}(D^3N^2)$, which is significantly faster than classical KMP.

\section{Conclusion}

We presented a formulation of kernelized movement primitives with a \textit{soft} null space projector (NS-KMP). Such projector allows a secondary goal to modulate a primary one, in this case a learned trajectory. This modulation occurs in proportion to the covariance in the data. We leveraged the theoretical properties of this projector, as well as its computational efficiency, in letter-writing and handover tasks to quickly discover demonstration-guided, collision-free motions. NS-KMP, as a non-parametric approach,  outperformed classical KMP in terms of computational complexity and performed competitively against the parametric approach ProMP. Moreover, we showed that NS-KMP provides fast adaptation in a fraction of a second in tasks where multi-dimensional input signals drive the robot behavior -- an important part of human-robot interaction.

In future work we will leverage the scalability of NS-KMP to solve adaptation problems in higher-dimensional settings, such as humanoid robot control. We will also investigate how to integrate more diversified types of multi-dimensional inputs into the NS-KMP framework such as EMG signals \cite{Vogel2020}. Finally, we will investigate similar null space projectors for Gaussian processes.

\section*{Acknowledgements}
This work has received funding from the European Union's Horizon 2020 Research and Innovation Programme under Grant Numbers 951992 (project VeriDream) and 101070600 (project SoftEnable), as well as under the Marie Skłodowska-Curie grant agreement No 101018395. We would also like to thank Alexander Dietrich for the discussions on the algebraic properties of null space projectors and Antonin Raffin for assisting with Optuna.

\bibliographystyle{./style/IEEEtran}

\end{document}